\documentclass[english,letterpaper, 10 pt, conference]{ieeeconf}
\usepackage[T1]{fontenc}
\usepackage[latin9]{inputenc}
\usepackage{babel}
\usepackage{array}
\usepackage{refstyle}
\usepackage{booktabs}
\usepackage{url}
\usepackage{multirow}
\usepackage{amstext}
\usepackage{graphicx}
\usepackage{rotating}
\usepackage{microtype}
\usepackage[unicode=true,pdfusetitle,
 bookmarks=true,bookmarksnumbered=false,bookmarksopen=false,
 breaklinks=false,pdfborder={0 0 1},backref=false,colorlinks=false]
 {hyperref}

\makeatletter

\AtBeginDocument{\providecommand\secref[1]{\ref{sec:#1}}}
\AtBeginDocument{\providecommand\figref[1]{\ref{fig:#1}}}
\AtBeginDocument{\providecommand\Figref[1]{\ref{Fig:#1}}}
\AtBeginDocument{\providecommand\appendixref[1]{\ref{appendix:#1}}}
\AtBeginDocument{\providecommand\Tabref[1]{\ref{Tab:#1}}}
\AtBeginDocument{\providecommand\subsecref[1]{\ref{subsec:#1}}}
\AtBeginDocument{\providecommand\tabref[1]{\ref{tab:#1}}}
\providecommand{\tabularnewline}{\\}
\RS@ifundefined{subsecref}
  {\newref{subsec}{name = \RSsectxt}}
  {}
\RS@ifundefined{thmref}
  {\def\RSthmtxt{theorem~}\newref{thm}{name = \RSthmtxt}}
  {}
\RS@ifundefined{lemref}
  {\def\RSlemtxt{lemma~}\newref{lem}{name = \RSlemtxt}}
  {}

\IEEEoverridecommandlockouts
\newref{sec}{%
name = \RSsectxt,
names = \RSsecstxt,
Name = \RSSectxt,
Names = \RSSecstxt,
refcmd = {\ref{#1}},  % remove the fucking \S!!!
rngtxt = \RSrngtxt,
lsttwotxt = \RSlsttwotxt,
lsttxt = \RSlsttxt}

\newref{subsec}{%
name = \RSsectxt,
names = \RSsecstxt,
Name = \RSSectxt,
Names = \RSSecstxt,
refcmd = {\ref{#1}},  % remove the fucking \S!!!
rngtxt = \RSrngtxt,
lsttwotxt = \RSlsttwotxt,
lsttxt = \RSlsttxt}

\newref{appendix}{%
name = \RSappendixname,
names = \RSappendicesname,
Name = \RSAppendixname,
Names = \RSAppendicesname,
refcmd = {\ref{#1}},  % remove the fucking \S!!!
rngtxt = \RSrngtxt,
lsttwotxt = \RSlsttwotxt,
lsttxt = \RSlsttxt}

\newcommand{\customfootnotetext}[2]{{% Group to localize change to footnote
  \renewcommand{\thefootnote}{#1}% Update footnote counter representation
  \footnotetext[0]{#2}}}% Print footnote text

\newcommand{\imageFootmark}{$^{\textrm{3}}$}
\newcommand{\imageFootnote}{
\customfootnotetext{3}{All images are brightness adjusted, 
cropped to the region of interest, and
best viewed when printed in color.}}

\usepackage{tikz}
\newcommand\copyrighttext{%
  \footnotesize \textcopyright\ 2021 IEEE\@.  Personal use of this material is permitted.  Permission from IEEE must be obtained for all other uses, in any current or future media, including reprinting/republishing this material for advertising or promotional purposes, creating new collective works, for resale or redistribution to servers or lists, or reuse of any copyrighted component of this work in other works.}
\newcommand\copyrightnotice{%
\begin{tikzpicture}[remember picture,overlay]
\node[anchor=south,yshift=10pt] at (current page.south) {\fbox{\parbox{\dimexpr\textwidth-\fboxsep-\fboxrule\relax}{\copyrighttext}}};
\end{tikzpicture}%
}

\makeatother

\begin{document}
\title{\Large \bf A Dataset for Provident Vehicle Detection at Night }
\author{Sascha Saralajew,$^{\textrm{*, 1}}$\thanks{$^{\textrm{*}}$Authors contributed equally.}\thanks{$^{\textrm{1}}$Bosch Center for Artificial Intelligence, Renningen,
and Leibniz University Hannover, Institute of Product Development,
Hannover, both Germany. The research was performed during employment
at Dr.\ Ing.\ h.c.\ F.\ Porsche\ AG\@. \texttt{sascha.saralajew@bosch.com}} Lars Ohnemus,$^{\textrm{*, 2}}$\thanks{$^{\textrm{2}}$Dr.\ Ing.\ h.c.\ F.\ Porsche AG, Weissach, Germany.
\texttt{\{lars.ohnemus1, lukas.ewecker, ebubekir.asan\}@porsche.de}} Lukas Ewecker,$^{\textrm{*, 2}}$ Ebubekir Asan,$^{\textrm{*, 2}}$
Simon Isele,$^{\textrm{2}}$ and Stefan Roos$^{\textrm{2}}$}
\maketitle
\begin{abstract}
In current object detection, algorithms require the object to be directly
visible in order to be detected. As humans, however, we intuitively
use visual cues caused by the respective object to already make assumptions
about its appearance. In the context of driving, such cues can be
shadows during the day and often light reflections at night. In this
paper, we study the problem of how to map this intuitive human behavior
to computer vision algorithms to detect oncoming vehicles at night
just from the light reflections they cause by their headlights. For
that, we present an extensive open-source dataset containing 59\,746%

annotated grayscale images out of 346%

different scenes in a rural environment at night. In these images,
all oncoming vehicles, their corresponding light objects (e.\,g.,
headlamps), and their respective light reflections (e.\,g., light
reflections on guardrails) are labeled. In this context, we discuss
the characteristics of the dataset and the challenges in objectively
describing visual cues such as light reflections. We provide different
metrics for different ways to approach the task and report the results
we achieved using state-of-the-art and custom object detection models
as a first benchmark. With that, we want to bring attention to a new
and so far neglected field in computer vision research, encourage
more researchers to tackle the problem, and thereby further close
the gap between human performance and computer vision systems.
\end{abstract}

\setcounter{footnote}{3}

\section{Introduction}

\copyrightnotice Object detection is one of the most popular and
best studied fields in computer vision. One reason is that object
detection is essential to several applications, for example, autonomous
driving and surveillance cameras. Currently, State-Of-The-Art (SOTA)
object detection is performed by Neural Networks (NNs) and the target
is to predict Bounding Boxes (BBs) that describe the position of \emph{visible}
objects. Thereby, a not negligible restriction is the visibility assumption
in form of direct sight to objects (ignoring systems that combine
detection and tracking to deal with occluded objects). In contrast,
in certain situations, human perception is capable to reason about
the position of objects without direct visibility---for instance,
by considering the illumination changes in an environment at daylight
(e.\,g., in form of shadow movements). When driving a car at night,
a similar human ability becomes a key feature for safe and anticipatory
driving: The ability to reliably detect light reflections in the environment
caused by the mandatory headlamps of other road participants. Building
on this ability, drivers can predict other vehicles and their position,
in the sense where the object will occur in the future, even though
they are \emph{not} in direct sight.

Motivated by the aforementioned human provident behavior and the discrepancy
to SOTA object detection, we provide with this paper all the tools
needed to boost ``ordinary'' vehicle detection at night to provident
vehicle detection such that consumer products can take advantage of
this ability soon.\imageFootnote  To this end, we introduce and
publish\footnote{\url{https://www.kaggle.com/saralajew/provident-vehicle-detection-at-night-pvdn}}
an annotated image dataset containing 346%

different sequences with 59\,746%

annotated grayscale images in total captured at two different exposures.
Each image is equipped with several annotations---in form of KeyPoints
(KPs)---that describe the position of other vehicles and all their
produced light instances (effects). Exemplary, we use this dataset
to train several machine learning models to detect light reflections
that might serve as baselines for future research. In order to properly
evaluate object detectors on the proposed dataset, we discuss commonly
used evaluation metrics in object detection and propose specially
crafted metrics for the dataset. Beyond the presented application
of provident vehicle detection at night, the structure of the dataset
is carefully designed such that it will be useful for other investigations
as well (e.\,g., object tracking at night, transfer learning between
different image exposure cycles).

\begin{figure}
\begin{centering}
\includegraphics[viewport=800bp 400bp 1260bp 600bp,clip,width=1\columnwidth]{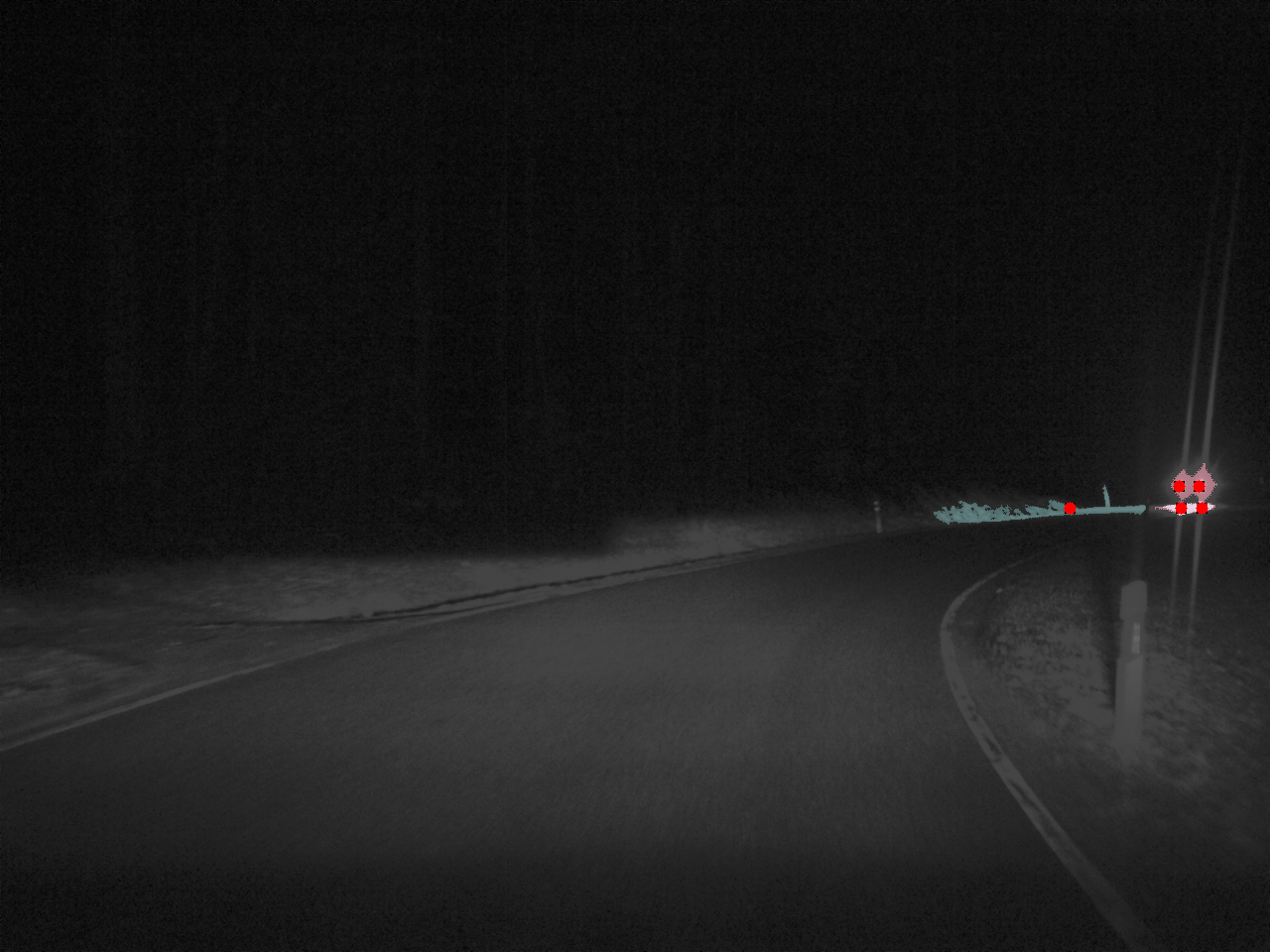}
\par\end{centering}
\caption{\label{fig:Salient-region-of}Annotation of light artifacts with KPs
(dots) and the corresponding inferred saliency maps (colored areas
represent the saliency of one KP).\imageFootmark}
\end{figure}
The outline of the paper is as follows: In the next section, we discuss
related work. Following this, we describe the annotation methodology
and the challenges in annotating light reflections. With these insights,
\secref{DATASET} explains the process of creating the Provident Vehicle
Detection at Night (PVDN) dataset. Afterwards, \secref{EXTENDING-THE-BASE}
shows the versatility of our annotation strategy by extending the
dataset with  saliency- and BB-representations. To analyze the performance
of detectors on the dataset, we craft evaluation metrics in \secref{Evaluation-metrics}.
After that, we present benchmark models trained on the dataset and
finish with a conclusion and discussion.
\begin{figure*}
\includegraphics[scale=0.23]{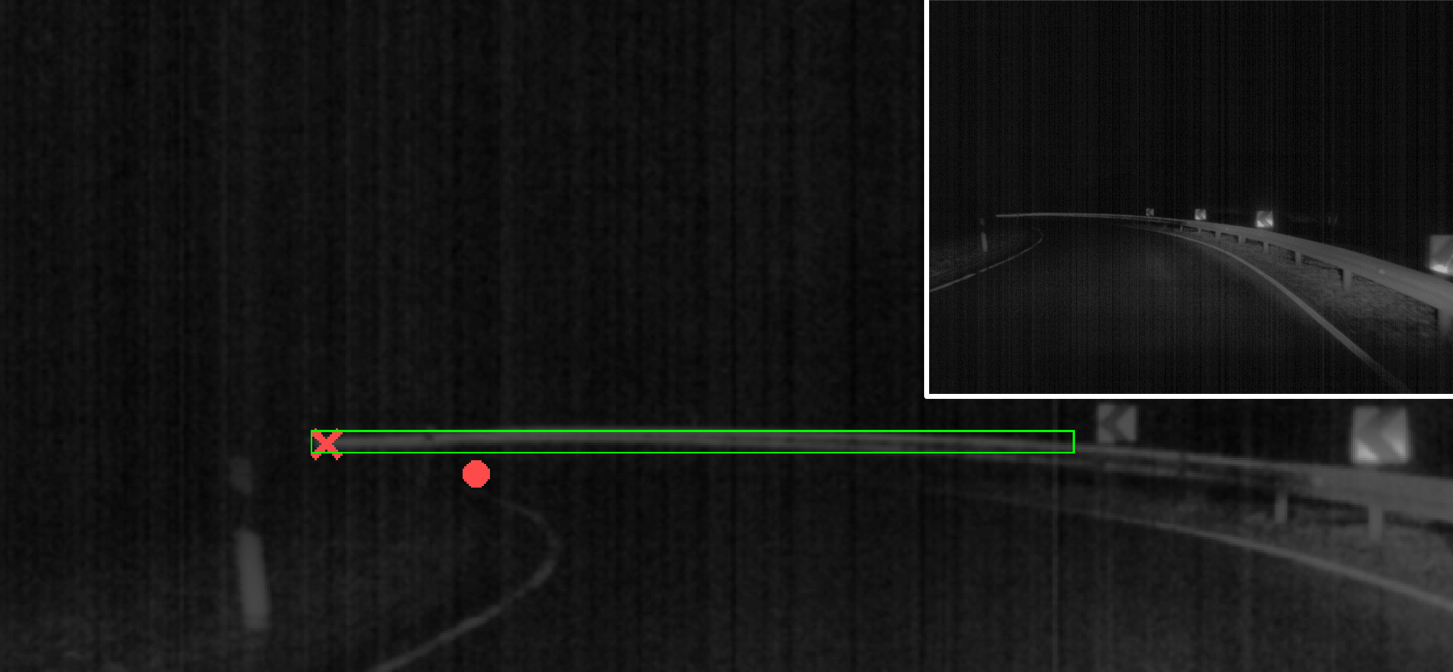}\hfill{}\includegraphics[scale=0.23]{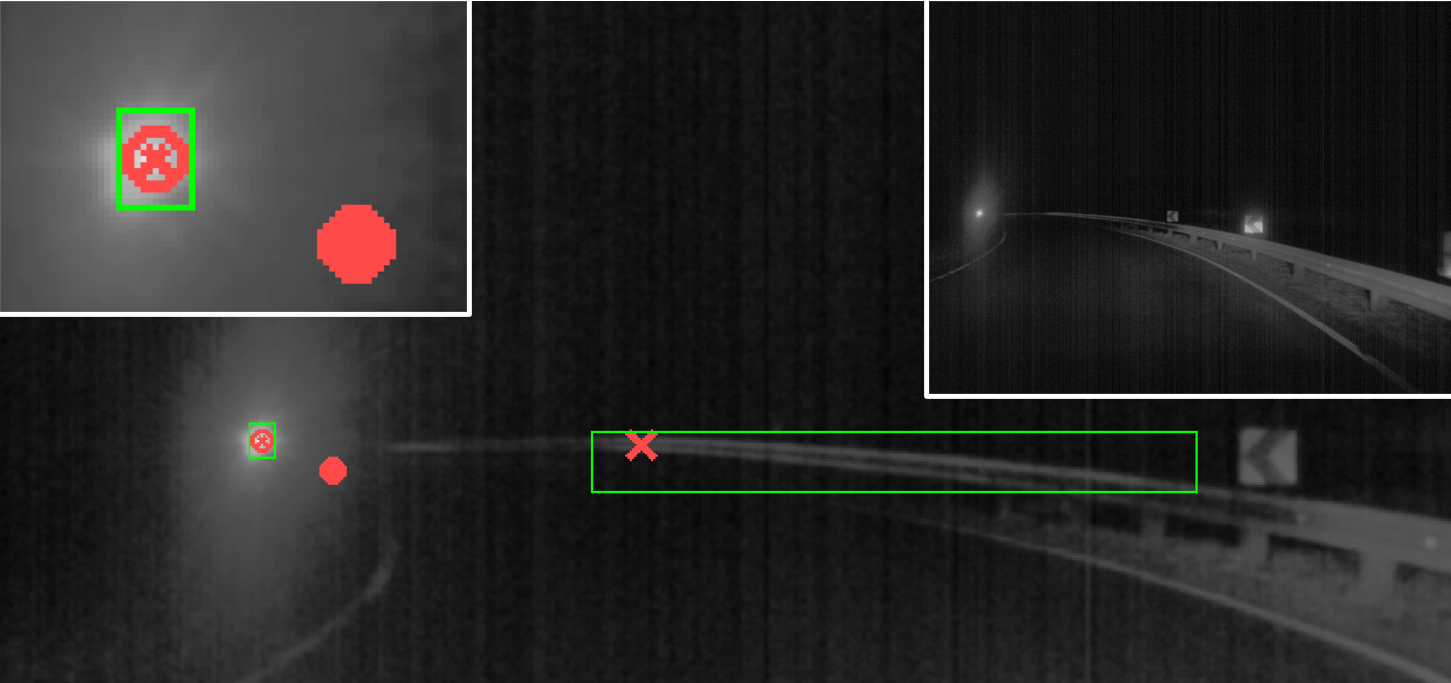}\hfill{}\includegraphics[scale=0.23]{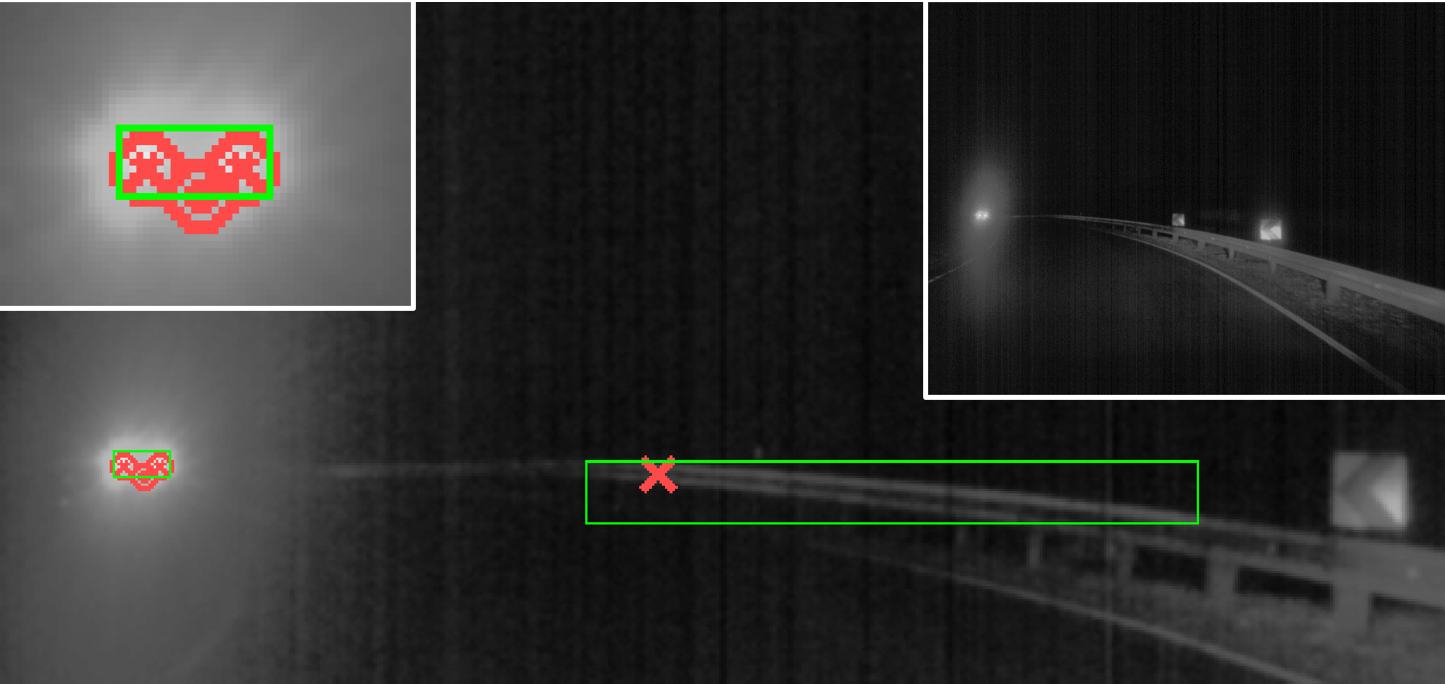}

\caption{\label{fig:opener-img}The three characteristic detection states in
an example scene (number 309) with the following annotations: indirect
vehicle position KPs (dot), direct vehicle position KPs (circle-dot),
indirect instance KPs (cross), direct instance KPs (circle-cross),
and automatically inferred BBs. Each visualization shows a cropped
area of the entire image (shown in the upper right corner). If necessary,
the left upper corner shows a second crop of the area of interest.
Left: First visible light reflections at the guardrail. Middle: First
direct sight to the vehicle. Right: Full direct sight to the vehicle.}
\end{figure*}

\section{Related Work\label{sec:Related-Work}}

Due to their broad application for driver assistance systems (e.\,g.,
automatic high beam control), the detection of vehicles at night is
well-studied. Current SOTA detectors detect vehicles by their visible
headlamps and rear lights~\cite{Lopez.2008,P.F.Alcantarilla.2011,Eum.2013,Juric.2014,Sevekar.2016,Satzoda2019}
so that these detection systems are limited to the detection of visible
vehicles solely. Therefore, published datasets alongside these papers
focus on the annotation of headlamps and rear lights (see also \cite{Rapson2018}).

Oldenziel~et~al.~\cite{Oldenziel2020} studied the deficit between
the human provident vehicle detection and a driver assistance system
camera-based vehicle detection system. Based on a test group study,
they showed that drivers detected oncoming vehicles on average 1.7\,s
before the camera system, which is a not negligible time discrepancy.
Later in their work, they used a Faster-RCNN architecture~\cite{fasterrcnn}
to detect light reflections caused by other vehicles. For this purpose,
they crafted a dataset (\textasciitilde 700 images) by annotating
light reflections and headlights with BBs. Even if the NN was able
to learn the task to some extent, they questioned whether BB annotations
are the correct annotation method since the absence of clear boundaries
of light reflections caused a high annotation uncertainty.\footnote{Note that this paper is a continuation of this earlier work by our
team.}

Orthogonal to the presented work here, Naser~et~al.~\cite{FelixMaximilianNaser.18.01.2019}
analyzed shadow movements to providently detect objects at daylight.
For that, they assumed that the ground plane is fixed and known to
analyze consecutive image frames by their homography. Due to this
assumption, their framework is not applicable to providently detect
vehicles at night.

\section{Annotation Methodology\label{sec:Annotation-Methodology}}

Usually, creating a dataset for object detection is straightforward:
Objects inside images are annotated by placing BBs around them. Therefore,
each BB directly specifies the image region that is taken up by an
object. Together with a class label, this perfectly specifies the
object. This, however, is not true for the task of provident detection
since, inherently, there is no image region associated with the object
we want to detect. Therefore, we have to deduce the existence of oncoming
cars by studying light artifacts in images. Unlike ordinary object
detection, light artifacts present several challenges due to their
fuzzy nature, abundance of light reflections, and glare. It gets even
more complicated when, in a realistic scene with multiple oncoming
vehicles and other light sources such as streetlights, we need to
match the light artifacts to their sources in order to infer correctly.
So, while the overall goal---detecting oncoming vehicles early---is
easy to state, the precise definition of the methodology to derive
a suitable dataset is more complicated.

\subsection{Annotation of Vehicle Positions}

The annotation of an approaching vehicle poses two challenges:
\begin{enumerate}
\item How to represent the position within a frame, and
\item where to place the vehicle position when it is not in direct sight? 
\end{enumerate}
When the vehicle is in direct sight, the most efficient way to annotate
the position is to place a KP at a remarkable point on the vehicle.
Placing a KP on the road centrally between both headlights perfectly
specifies the vehicle's position---see the right image in \figref{opener-img}.
The choice of assigning a BB to the vehicle might seem to be more
natural but can be made redundant when the headlights of the vehicle
are marked by KPs as well. A robust BB can then be always inferred
from the vehicle position and the two headlight KPs by taking the
enveloping rectangle of all three KPs.

Stringently, the predicted position of not yet visible vehicles---we
will call them \emph{indirectly} visible---can be annotated with
a KP as well. With possible prediction algorithms in mind, this KP
is placed at the location of the first sight of the approaching vehicle:
the end of the visible road (if the road touches the edge of the image,
the intersection point is used)---see \figref{opener-img} (left
and middle) for an illustration. This approach has the key advantages
that it conserves temporal coherence when transitioning between indirect
and direct vehicle positions; and that the KP can be directly used
as a target when predicting the occurrence position of vehicles.

Future systems might need to make different decisions depending on
whether it is a directly or indirectly visible vehicle. To ease this,
a Boolean label is added to each vehicle position to specify whether
it is directly or indirectly visible.

\subsection{Notations and Annotation Hierarchy }

Generally, for annotation, we study image sequences (scenes) showing
the scenario ``approaching vehicle at night''. Within such scenes,
vehicles are detected. Since the \emph{vehicle position} is also an
abstract notion for not directly visible vehicles, the presence of
vehicles might be predicted through light artifacts within a frame.
To emphasize this relationship, we denote any form of light artifacts
as an \emph{instance} of the vehicle.
\begin{figure}
\begin{centering}
\includegraphics[width=0.8\columnwidth]{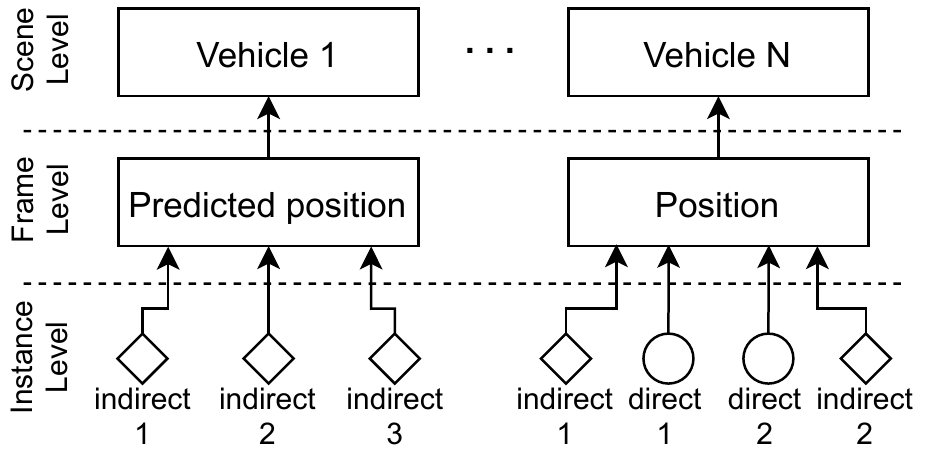}
\par\end{centering}
\caption{\label{fig:Hierachy-of-different}Hierarchy of different entities
used for the PVDN dataset. The structure allows for multiple vehicles
each having a single (predicted) position. The vehicles are observed
through a variable number of instances.}
\end{figure}
The ``human'' approach to predict and detect vehicles at night spans
three hierarchical semantic levels. On the top level\emph{---the
scene level}---the aim is to predict and locate oncoming vehicles.
This usually uses both temporal and spatial features. While the consideration
of consecutive frames allows the driver to increase the detection
certainty, oncoming vehicles can be detected using isolated, single
images only. On this \emph{frame level}, the vehicles have to be located.
As instances build up the knowledge about vehicles, they are placed
on the lowest semantic level, the \emph{instance level}. The instances
are essential because, for not directly visible vehicles, they are
the only features available to infer the vehicles' positions. \Figref{Hierachy-of-different}
shows the hierarchy that implies two requirements for a suitable annotation
strategy: 
\begin{enumerate}
\item vehicle positions and their instances must be annotated;
\item annotations should be coherent over multiple frames to capture the
temporal characteristics.
\end{enumerate}
While the second requirement can be enforced by assigning IDs to each
annotated entity, the annotation of vehicle positions and instances
requires further investigations.

\subsection{Annotation of Instances}

We split the annotations in the same way we did for the vehicle position.
Instances can be both direct (headlights) or indirect (halos, reflections),
see \figref{opener-img}. While direct instances are well-specified,
the total number and size of the indirect instances cannot be specified
objectively. The headlights might bounce and be reflected multiple
times, leaving instances of various strengths all around the image. 

\subsubsection{Lessons learned of BB annotations}

Oldenziel~et~al.~\cite{Oldenziel2020} used BBs to represent light
reflections, as already stated in \secref{Related-Work}, and noticed
a strong tendency of the detector towards False Positive (FP) detections.
We hypothesize, that this was mainly caused by two reasons: First,
only the strongest light artifact was annotated to represent a vehicle.
Even if this assumption resulted in an objective annotation strategy,
it seemed to confuse the system as the intensity of the strongest
instance is always dependent on the entire scenario. Thus, a lot of
predictions might be True Positives (TPs) but were considered FPs
due to the missing annotations of weaker instances. Second, the metric
used for evaluating predictions---the Intersection over Union (IoU)---is
sensible to size variations between ground truth and predictions.
While this is usually desired for object detection, it requires a
well-specified, objective ground truth. However, on many occasions,
both the ground truth and the prediction could be considered correct
while still giving a small IoU. This poses the question: Is it possible
to objectively specify BBs? To address this, we performed a small
experiment, tasking multiple experts to annotate the same indirect
instance with BBs. The results show that on average the experts' BBs
only overlapped with a median IoU of 59.4\%%
for the indirect instance. This effect is even worse for direct instances,
resulting in a median IoU of just 52.0\%%
.
For a more detailed discussion of the annotation experiment, see \appendixref{Studying-Annotation-Uncertainty}. 

\subsubsection{KP annotations based on saliency}

Based on the previous discussion, it is clear that a proper annotation
strategy should not insinuate the existence of perfect annotations
and therefore leading to false conclusions on the real problem. Thus,
a suitable strategy has to be flexible enough to allow for variations
while still describing the situation properly. Simply put, the annotation
strategy should be guided by the local light intensity because this
is the only available feature to judge the existence of instances.
In real life, such local light intensities or locally \emph{salient
regions} attract the attention of drivers, and the goal of the annotation
process is to mimic this attention behavior. 

An approach to this attention behavior is to mark points of high visual
saliency by KPs without proclaiming a boundary, see \figref{opener-img}.
With these KPs carefully placed in salient regions, we have markers
for human attention---similar to datasets using eye-fixation points
as descriptors of visual saliency~\cite{borji2015cat2000}. This
annotation approach captures the top-down aspect of human attention~\cite{itti2015computational},
which is goal-driven (detect oncoming vehicles). After the annotation,
we can use the KPs to check for intensity stimuli that lead the annotators
to the decision to place a KP in this region, which corresponds to
a bottom-up or stimuli-based saliency approach~\cite{itti2015computational}.
Such a bottom-up saliency approach can be realized by comparing the
pixel intensities of the surrounding area with the intensity value
at the KP\@. Usually, the intensity values in the surrounding area
are on average slightly lower than the KP intensity because the annotators
intuitively look at the most salient point first---the intensity
maximum. In \Figref{Salient-region-of} we show an example where the
salient region was inferred from the KP (see \secref{EXTENDING-THE-BASE}
for further details about the saliency inference). Using this approach,
we can specify entire instances without stating ground-truth borders
but rather suggestions of interesting image regions. Additionally,
the approach is fairly robust to variations in the KP placement. 

This annotation strategy applies to direct and indirect instances.
For direct instances, the KP is placed centrally within the headlight
cone, see \figref{opener-img}. To distinguish between both instance
types, we add a Boolean label. Additionally, it is worth noting, that
this strategy is efficient to carry out since only one point per instance
has to be annotated.

\section{The PVDN Dataset\label{sec:DATASET}}

\begin{table}
\centering{}\caption{\label{tab:Dataset-split}Training-validation-test split of the PVDN
dataset.}
\begin{tabular}{cccccc}
\toprule 
 & exposure & \#scenes & \#images & \#veh.-pos. & \#instances\tabularnewline
\midrule
\midrule 
\multirow{2}{*}{\begin{turn}{90}
train~\,
\end{turn}} & day & 113 & 19\,078 & 15\,403 & 45\,765\tabularnewline
\cmidrule{2-6} \cmidrule{3-6} \cmidrule{4-6} \cmidrule{5-6} \cmidrule{6-6} 
 & night & 145 & 25\,264 & 26\,615 & 72\,304\tabularnewline
\midrule 
\multirow{2}{*}{\begin{turn}{90}
val~~\,
\end{turn}} & day & 20 & 3\,898 & 2\,602 & 7\,244\tabularnewline
\cmidrule{2-6} \cmidrule{3-6} \cmidrule{4-6} \cmidrule{5-6} \cmidrule{6-6} 
 & night & 25 & 4\,322 & 3\,600 & 12\,746\tabularnewline
\midrule 
\multirow{2}{*}{\begin{turn}{90}
test~~
\end{turn}} & day & 19 & 3\,132 & 3\,045 & 9\,338\tabularnewline
\cmidrule{2-6} \cmidrule{3-6} \cmidrule{4-6} \cmidrule{5-6} \cmidrule{6-6} 
 & night & 24 & 4\,052 & 3\,384 & 10\,438\tabularnewline
\midrule 
\multirow{3}{*}{\begin{turn}{90}
total~~~~
\end{turn}} & day & 152 & 26\,108 & 21\,050 & 62\,347\tabularnewline
\cmidrule{2-6} \cmidrule{3-6} \cmidrule{4-6} \cmidrule{5-6} \cmidrule{6-6} 
 & night & 194 & 33\,638 & 33\,599 & 95\,488\tabularnewline
\cmidrule{2-6} \cmidrule{3-6} \cmidrule{4-6} \cmidrule{5-6} \cmidrule{6-6} 
 & cumulated & 346 & 59\,746 & 54\,649 & 157\,835\tabularnewline
\bottomrule
\end{tabular}
\end{table}
The PVDN dataset is derived from a test group study performed to analyze
provident vehicle detection skills of humans \cite{Oldenziel2020}.
During the study, the onboard camera of the test car was used to capture
grayscale images of two different exposure cycles (each at 18\,Hz).
The resulting image cycles are called ``day cycle'' (short exposure)
and ``night cycle'' (long exposure). Image sequences (scenes, available
through the test group study) are selected according to the appearance
of oncoming vehicles and padded to create parts without vehicles in
them. Mostly, such sequences begin when there is no sign of the oncoming
vehicle and end when the vehicle has passed. For each sequence either
the day or the night cycle is annotated according to the methodology
described in \secref{Annotation-Methodology}.

As already mentioned, the annotation of indirect instances and vehicle
positions is subjective. To minimize the subjectivity effects, an
iterative procedure for the annotation process was chosen. KPs for
vehicle position and instances were annotated with a custom annotation
tool.\footnote{\label{fn:PVDN-Github}\url{https://github.com/larsOhne/pvdn}}
Then, the annotations were reviewed by multiple annotators and the
placement and number of KPs was discussed. Additionally, a set of
guidelines for different conditions and circumstances was given to
the annotators. The reviewed annotations were then corrected by an
annotator.

The dataset contains 59\,746%

annotated images, spread over 346%

scenes. In comparison to the Enz-dataset \cite{Oldenziel2020}, the
share of indirect instances to direct instances is balanced (ca. $49\%$
of instances are indirect). The same is true when considering vehicle
positions ($43\%$ of the annotated vehicle positions are indirect).
The total numbers of scenes, images, vehicle positions, and instances
are summarized in \Tabref{Dataset-split}.

\section{Extending the Instance Annotations\label{sec:EXTENDING-THE-BASE}}

While KP annotations allow for a compact and natural annotation strategy,
they are not commonly used in object detection. Therefore, we provide
automated approaches to regress salient regions from the instance
KPs, representing them with both saliency maps and BBs. 

\subsection{Generation of Saliency Maps\label{subsec:Saliency-Maps}}

We use a modified version of the Boolean Map Saliency (BMS) method
introduced by Zhang and Sclaroff~\cite{zhang2013} to automatically
generate saliency maps. In the original approach, saliency was deduced
by thresholding image channels. The resulting Boolean maps are then
filtered using the assumption that salient regions are not connected
to the image borders (center-surround antagonism) and averaged to
retrieve attention maps. However, the center surround-antagonism does
not hold for the PVDN dataset because the instances are in general
not centered within the image. Thus, we replaced the original criterion
with a KP-centered approach. 

To retrieve saliency maps for the PVDN dataset, we apply a flood fill
algorithm with the KPs as seed pixels for the filtering in the Boolean
maps. Additionally, we improve the original BMS method by restricting
the maximal threshold value to approximately 1.2 times the KP intensity
because otherwise, the salient region might leak into other high-intensity
regions. In \figref{Salient-region-of}, we show an exemplary salient
region around a KP generated with this modified BMS method. 

Given a KP $\tilde{\mathbf{p}}$, the corresponding saliency map $s\left(\mathbf{p}\mid\tilde{\mathbf{p}}\right)$
assigns each pixel $\mathbf{p}\in I$ in the image $I$ a saliency
value $s:I\rightarrow[0,1]$. The saliency maps across all KPs can
be used to directly learn end-to-end saliency-based object detectors
(e.\,g., using methods described by Borji~et~al.~\cite{borji2015})
or can be used to improve KP metrics, see \secref{Evaluation-metrics}. 

\subsection{Generation of Bounding Boxes\label{subsec:Bounding-Boxes}}

There are the following two options to generate BBs for the PVDN dataset:
\begin{itemize}
\item use the KPs to filter predictions of a BB regressor;
\item regress BBs directly from the local intensities and KPs.
\end{itemize}
For filtering, we assume that a suitable system generated a list of
candidate BBs. Then, we validate each of these candidate BBs by checking
whether at least one KP is contained inside the BB\@. If there is,
we consider the BB a TP\@. Otherwise, the BB is considered a true
negative. 

One possible candidate generation we will use later in the paper is
based on the adaptive binarization approach proposed by Singh~et~al.~\cite{singh2011thresholding}.
This method efficiently determines binary masks of images where blobs
are highlighted by computing a locally adaptive binarization threshold
based on local means and mean deviations. Each blob is then converted
into a BB by computing the enclosing rectangle of the connected region.
In \figref{blob_detector}, the first three steps visualize the candidate
generation and \figref{opener-img} shows example TP BBs.

The other approach to generate BBs is similar to the saliency approach
and uses simple thresholding around the KP intensities to generate
a Boolean map. In this Boolean map, we filter the instance blobs by
applying a blob detection algorithm with the KPs as seeds. Analogously,
binarization thresholds can be chosen relative to the KP intensities,
which makes the BBs more robust for weak instances. A more detailed
analysis of the BB generation is provided in \cite{ewecker2021provident}.

\section{Evaluation Metrics\label{sec:Evaluation-metrics}}

To properly evaluate object detection methods on the PVDN dataset,
we provide specially crafted evaluation metrics for BB and KP predictors. 

\subsection{Evaluation Metric for Bounding Box Predictors\label{subsec:Evaluation-Metric-BB}}

As explained in \subsecref{Bounding-Boxes}, it is possible to infer
BBs from the KP annotations. To enable the use of current SOTA object
detection methods---that usually predict BBs---on the PVDN dataset,
we provide a metric to evaluate their performance based on the ground
truth KPs. Without loss of generality,\footnote{The evaluation can be extended to multiple classes (e.\,g., direct
instance, indirect instance, and no instance). However, defining it
as a binary problem avoids the task to resolve ambiguities (when a
BB contains a direct KP and an indirect KP).} we define the evaluation method for binary BB predictors. Hence,
the task is to determine whether a BB corresponds to an instance (either
direct or indirect).

When detecting and classifying objects via BBs, the most common metrics
to evaluate model performance are mean Average Precision (mAP) and
mean Average Recall (mAR). For both, the match between a ground truth
BB and a predicted BB is determined by calculating the Intersection
over Union (IoU). If the IoU between a ground truth and predicted
BB exceeds a certain threshold, both are considered a match and the
classification performance can be determined by comparing the ground
truth and predicted label. Obviously, determining the correspondence
of ground truth and predicted BBs via IoU is only feasible if the
ground truth annotations provide BBs. Since the instances in the PVDN
dataset are annotated via KPs, calculating the IoU is not possible.
In this case, our proposed metric to compare predicted BBs with ground
truth KPs follows the idea that 
\begin{itemize}
\item each BB should span around exactly one KP, and
\item each KP should lie within exactly one BB\@.
\end{itemize}
Consequently, it should be quantified when \textit{a)} a BB contains
no KP, \textit{b)} a BB contains more than one KP, \textit{c)} a KP
does not lie within any BB, and \textit{d)} a KP lies within more
than one BB\@.

First, to compute precision and recall in the usual manner, to quantify
\emph{a)} and \emph{c)}, we introduce the following events:
\begin{itemize}
\item True Positive (TP)\@: The  instance is described (at least one BB
spans over the KP);
\item False Positive (FP)\@: The BB describes no instance (no KP lies within
the BB);
\item False Negative (FN)\@: The  instance is not described (no BB spans
over the KP).
\end{itemize}
Second, to quantify the BB quality of TP events, we calculate the
following quantities:
\begin{itemize}
\item The number of KPs in a BB $b$ that is a TP, denoted by $n_{K}\left(b\right)$;
\item The number of BBs that are TPs and that cover a described KP $k$,
denoted by $n_{B}\left(k\right)$.
\end{itemize}
These numbers are converted into measures in the range $\left[0,1\right]$,
where one means best quality, and averaged across the dataset by
\[
q_{K}=\frac{1}{N_{B}}\sum_{b}\frac{1}{n_{K}\left(b\right)}
\]
and
\[
q_{B}=\frac{1}{N_{K}}\sum_{k}\frac{1}{n_{B}\left(k\right)}.
\]
Thereby, $N_{B}$ is the total number of BBs that are TPs and $N_{K}$
is the total number of described KPs. These measures quantify the
uniqueness of BBs with respect to \emph{b)} and \emph{d)}. For example,
in an image with several KPs, $q_{k}$ is low if a large BB spans
over the whole image, since it will then capture several KPs and the
term $\frac{1}{n_{K}\left(b\right)}$ decreases. The overall quality
is determined by $q=q_{K}\cdot q_{B}$.

\subsection{Evaluation Metric for Keypoint Predictors}

Instead of predicting BBs, another option to solve the problem is
to predict KP coordinates for each  instance. Usually, KP prediction
tasks evaluate methods also by calculating mAP and mAR but determining
the correspondence of ground truth to prediction not via IoU but a
KP similarity measure. Inspired by the ``object keypoint similarity''
method\footnote{\url{https://cocodataset.org/\#keypoints-eval}} proposed
in the COCO~\cite{coco} KP detection challenge, we propose a way
to measure KP similarity, denoted by $\mathrm{kps}\left(\cdot,\cdot\right)$.
In particular, for the PVDN dataset, we measure the similarity between
a ground truth KP and a predicted KP by 
\[
\mathrm{kps}\left(\mathbf{p},\tilde{\mathbf{p}}\right)=\exp\left(-\left\Vert \mathbf{p}-\tilde{\mathbf{p}}\right\Vert _{E}\cdot\left|s\left(\mathbf{p}\mid\tilde{\mathbf{p}}\right)-s\left(\tilde{\mathbf{p}}\mid\tilde{\mathbf{p}}\right)\right|\right),
\]
where $\tilde{\mathbf{p}}$ refers to the pixel coordinate vector
of a ground truth KP and $\mathbf{p}$ refers to the pixel coordinate
vector of a predicted KP\@. The function $s\left(\mathbf{p}\mid\tilde{\mathbf{p}}\right)$
returns the saliency at the position $\mathbf{p}$ with respect to
the saliency map of the ground truth KP $\tilde{\mathbf{p}}$, see
\subsecref{Saliency-Maps}. With this measure, the ambiguity of the
ground truth KPs is addressed by scaling the Euclidean distance between
the predicted and ground truth KP by the difference of the saliencies
at both positions. Following the idea that an instance is annotated
with a KP placed on its intensity---and therefore saliency---maximum
objectively, the Euclidean distance between ground truth and predicted
KP is scaled down the more similar the saliency is. This takes into
account the noisy nature of the ground truth. Once the similarity
score is calculated for each combination of predicted KP and ground
truth KP, the optimal assignment between the prediction and ground
truth is calculated by formulating the problem as finding a bipartite
matching with a maximal similarity score. This problem can be efficiently
solved by the Hungarian method. Having this optimal assignment, mAP
and mAR can be calculated as usual.

\section{Benchmark Experiments\label{sec:Benchmark-experiments}}

\begin{figure}
\hfill{}\includegraphics[width=1\columnwidth]{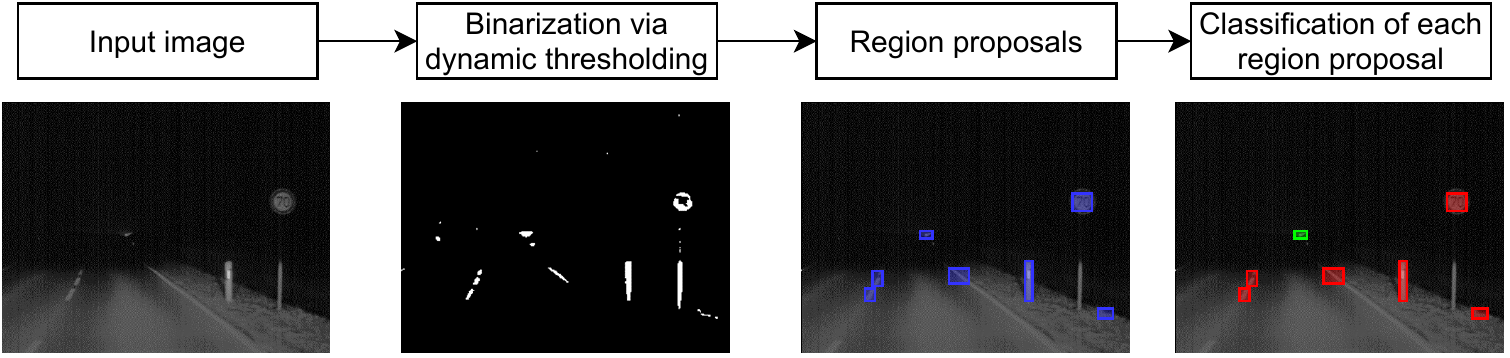}\hfill{}
\raggedright{}\caption{\label{fig:blob_detector}Architecture overview of the custom model.}
\end{figure}
\begin{table*}
\centering{}\caption{\label{tab:Model-results}Model results on the PVDN dataset determined
by the metrics proposed in \secref{Evaluation-metrics} on the day-test
dataset.}
\begin{tabular}{ccccccccccc}
\toprule 
Model & Input size & Parameters {[}M{]} & GFLOPs & Runtime {[}ms{]} & Precision & Recall & F-score & $q$ & $q_{K}$ & $q_{B}$\tabularnewline
\midrule
YoloV5x & 960$\thinspace\times\thinspace$960 & 87.2 & 217.1 & 28.1 & 1.00 & 0.68 & 0.81 & 0.37 & 0.38$\thinspace\pm\thinspace$0.20 & 0.98$\thinspace\pm\thinspace$0.08\tabularnewline
YoloV5s & 960$\thinspace\times\thinspace$960 & 7.1 & 16.4 & 14.8 & 0.99 & 0.66 & 0.80 & 0.37 & 0.38$\thinspace\pm\thinspace$0.20 & 0.98$\thinspace\pm\thinspace$0.09\tabularnewline
Custom & 480$\thinspace\times\thinspace$640 & 0.9 & 1.6$\thinspace\pm\thinspace$0.7 & 20.6$\thinspace\pm\,$3.5 & 0.88 & 0.54 & 0.67 & 0.40 & 0.40$\thinspace\pm\thinspace$0.22 & 1.00$\thinspace\pm\thinspace$0.00\tabularnewline
\multicolumn{5}{c}{Generated BB day-test dataset} & 1.00 & 0.69 & 0.81 & 0.42 & 0.42$\thinspace\pm\thinspace$0.24 & 1.00$\thinspace\pm\thinspace$0.00\tabularnewline
\bottomrule
\end{tabular}
\end{table*}
SOTA object detection methods usually predict BBs. Although the dataset
is annotated with KPs, our benchmark studies at this point focus on
predicting BBs for two reasons. First, we want to show that based
on the KP annotations the derived BBs, see \subsecref{Bounding-Boxes},
can serve as a useful training basis. Second, SOTA KP prediction methods
(e.\,g.,~\cite{wei2016posemachines,sun2019deep,zhang2020distribution})
cannot be applied out of the box without greater architecture adjustments
because of
\begin{itemize}
\item the inherently noisy nature of the instance annotations via KPs, and
\item the non-fixed number of KPs per vehicle and frame.
\end{itemize}
In a proof-of-concept, we investigated the applicability of KP predictors.
However, the results were not satisfying and far behind the results
of BB methods. Therefore, we decided to focus on BB-based object detection
in this study. The source code of the experiments is publicly available.\fnref*{PVDN-Github}

\subsection{Model architectures}

We evaluated three models: YoloV5x, its lightweight version YoloV5s,\footnote{\url{https://github.com/ultralytics/yolov5}}
and a customized approach, denoted as Custom. A more detailed analysis
of this approach is presented by Ewecker et al. \cite{ewecker2021provident}.
The YoloV5 architecture was formalized to predict instance BBs (either
direct or indirect). The Custom architecture is a combination of a
rule-based region proposal module and a Convolutional NN (CNN) classifier---see
\figref{blob_detector} for the architecture overview and \tabref{Classifier-architecture-Custom}
in the appendix for the layer structure of the CNN\@. In the first
step of the Custom architecture, we generate BB candidates (see the
next section for details). These generated BBs, are used to crop image
patches from the input image. Thereby, the cropped region has twice
the size of the given BB in order to include additional context information.
In the second step, the image patches are resized to 64$\thinspace\times\thinspace$64
pixels and fed into the binary CNN classifier. The task of this classifier
was to assign the label positive (instance) or negative to the BBs.

\subsection{Evaluation Setup and Training}

All evaluations are implemented in Python and C++. The NNs are implemented
in PyTorch~\cite{NEURIPS2019_9015} and are executed on an NVIDIA
Tesla V100 GPU\@. The main Python process and the C++ parts are running
on a Intel Core i7-6850K 3.60GHz CPU\@.

We took the \emph{day exposure} subset and followed the proposed adaptive
thresholding method of \subsecref{Bounding-Boxes} to generate the
BBs. For this purpose, we resized the images to half the size and
tuned all available hyperparameters by a random search on the validation
split. After that, we used the BB regressor also as the region proposal
method for the Custom architecture. All the NNs were trained on the
training split for at most 500 epochs until convergence. The final
models were then selected based on the maximum F-score on the validation
set over all epochs and results are reported on the test set. These
evaluations were done with respect to the evaluation measures proposed
in \subsecref{Evaluation-Metric-BB}. Besides model performance, we
also report the number of parameters and the computational complexity
in terms of Giga Floating Point Operations (GFLOPs) per image as a
hardware-independent measure. Additionally, model runtimes are given.
If available, we report standard deviations (mean\,$\pm$\,std).

\subsection{Results}

The numerical results are summarized in \tabref{Model-results}. At
first, we evaluated the quality of the generated ground truth BBs.
Note that, in the best case, all of the performance metrics give a
score of 1.0. As expected, the generated BBs yield a perfect precision
of 1.0 as they naturally do not contain any FPs. However, they only
provide a recall of 0.69 indicating that several KP annotations do
not correspond to any BB (FN). The analysis of these errors shows
that they are caused when the annotated instances have a very low
saliency such that they are ignored by the adaptive thresholding.
Furthermore, a box quality of 0.42 reveals that the applied method
of generating BBs based on the KP annotations still has difficulties
in separating instances from each other. This means that when instances
are close to each other the intensity difference is not sufficient
for the adaptive thresholding to distinguish the blobs and, therefore,
to create object boundaries. In such cases, instances, which are annotated
separately, are being described by one BB (e.\,g., see the right
image in \figref{opener-img}).

The best performance in terms of the F-score is achieved by the YoloV5
versions and these results are close to the scores of the generated
BBs. Of course, as the learning scheme is supervised, the scores should
not exceed the scores evaluated on the training data. With respect
to the recall and the bounding box quality, the scores suggest that
the YoloV5 was able to learn or approximate the BB generation approach
that we have used for the dataset generation, which is a remarkable
result. Behind the scores of the YoloV5 versions is the Custom approach
with a non-trivial result, which shows that it is possible to solve
this task by a combination of a rule-based and learned approach. 

In terms of memory and computational complexity, the Custom approach
is significantly superior to the YoloV5 versions. However, in contrast
to the computational complexity, the Custom method is not faster (runtime)
than the YoloV5s. The reason for that is that the C++ implementation
of the BB generation is not highly optimized. To put this into numbers,
from the total computation time on average only 2.5\,ms are required
to compute the classification outputs of the CNN per image (including
data loading to the GPU).\footnote{Note that the computational complexity of the Custom approach is not
constant as the runtime of the proposal generation depends on the
number of positive values in the Boolean map.}

Overall, our results show that \textit{a)} it is possible to detect
oncoming vehicles based on their light reflections and \textit{b)}
based on the KP annotations BBs can be generated that can serve as
training data. 

Finally, a rationale for the design of the Custom approach: We used
the Custom method together with a tracking and plausibility algorithm
to test prototype driver assistance functions in a test car. For that,
we ran the Custom approach together with all the other applications
(receiving inputs, sending outputs, etc.) in real time (faster than
18\,Hz). The usage of a shallow CNN together with a rule-based BB
generator has the advantage to be more transparent. Therefore, the
model behavior and, especially, exceptional cases can be better understood,
analyzed, and interpreted making the approach easier to validate and
verify.

\section{Conclusion and Outlook}

Based on the work of Oldenziel~et~al.~\cite{Oldenziel2020}, we
presented a thoroughly designed image dataset including evaluation
metrics and benchmark results to investigate the problem of provident
vehicle detection at night. In general, the task to providently detect
objects based on visual cues is a promising new research direction
to bring object detection towards human performance. With the presented
work here and the open-sourcing of all data and code, we hope to provide
the research community a jump start.

The annotation of light artifacts is challenging compared to ordinary
object annotation due to the lack of clear object boundaries. We tackled
this difficulty by an annotation approach based on placing KPs in
salient regions and methods to automatically extend the base annotations
to saliency maps or BBs. Even if we showed in a set of benchmark experiments
that automatically generated BBs can be used to train SOTA object
detectors such that they detect light reflections, the generation
approach and their inherent quality is of course a limiting factor.
Since we already provided an evaluation metric for KP detectors, we
plan to develop KP and saliency detection methods to overcome this
limiting factor.

Besides, we annotated the vehicle positions so that the PVDN dataset
provides ground truth for several research fields: 
\begin{itemize}
\item Time estimation: Time until vehicle appearance;
\item Tracking: Track instances and vehicles across scenes;
\item Transfer learning: Transfer methods to different exposure cycles or
weather conditions (snow, rain, fog, etc.);
\item Learning with label noise: Design methods that properly handle the
KP or generated annotations.
\end{itemize}
We are looking forward to all the upcoming results.

\bibliographystyle{IEEEtran}
\bibliography{paper}

\appendices

\section{Bounding Box Annotation Consistency\label{appendix:Studying-Annotation-Uncertainty}}

Usually, BB annotations created by experts for object detection can
be considered as objective ground truth. However, for instances, this
ground truth is subjective. To study this subjectiveness, we compared
the BB annotations for direct and indirect instances across several
annotators. The results confirm the following hypothesis: \emph{BB
annotations vary to a not negligible extent between different annotators
or even for the same annotator between different frames}.

\subsection{Experiment Setup}

To study the annotation consistency, seven%
experts were tasked to annotate the same scene with BBs. Thereby,
the annotation inconsistency that might be observed is two-fold: First,
some experts might decide to not annotate an instance because it is
too weak or cannot be separated from another instance. Second, after
the experts agreed on the existence of an instance, the actual size
and position of the BB are also subject to the subjective behavior
of the annotators. 

We decided to only study the subjectivity of the BB placement. This
decision was made to reduce the evaluation complexity of the experiment
while still emphasizing the overall problem. This led to the following
experiment setup:
\begin{enumerate}
\item Two instances (one direct, one indirect) were selected for annotation.
The participants were shown an example image to clarify the annotation
targets.
\item The experts annotate the scene (number 15) from start to end using
BBs.
\item The resulting annotations are evaluated with appropriate metrics.
\end{enumerate}

\subsection{Evaluation Metrics}

For evaluation, annotated BBs were collected from all $N$ experts.
To explain the evaluation scheme, we will focus on the consistency
of all BBs $B=\left\{ b_{1},\dots,b_{N}\right\} =\left\{ \left(\mathbf{p}_{1,1},\mathbf{p}_{1,2}\right),\dots,\left(\mathbf{p}_{N,1},\mathbf{p}_{N,2}\right)\right\} $
for an instance in one frame. Metrics are calculated for the direct
and indirect instances separately and then averaged over all frames.
A BB is described by its upper-left corner $\mathbf{p}_{i,1}$ and
its lower-right corner $\mathbf{p}_{i,2}$. $\mathrm{Rect}\left(b_{i}\right)$
defines the set of all points inside the BB $b_{i}$. Drawing from
common object detection metrics, the IoU can be used to compress the
position and size variations of BBs into a scalar value. This is usually
done by comparing a predicted BB to a ground truth BB\@. Since this
is not applicable for our purpose, we compare the expert annotations
to the mean BB $\bar{b}$, specified as the average over all annotations:
\[
\bar{b}=\left(\mathbf{\bar{p}}_{1},\mathbf{\bar{p}}_{2}\right)=\left(\frac{1}{N}\sum_{i=1}^{N}\mathbf{p}_{i,1},\frac{1}{N}\sum_{i=1}^{N}\mathbf{p}_{i,2}\right).
\]
Using this as a reference, the IoU values to all BBs are calculated:
\[
\mathrm{IoU}\left(b_{i},\bar{b}\right)=\frac{\left|\mathrm{Rect}\left(b_{i}\right)\cap\mathrm{Rect}\left(\bar{b}\right)\right|}{\left|\mathrm{Rect}\left(b_{i}\right)\cup\mathrm{Rect}\left(\bar{b}\right)\right|}.
\]
Assuming, that all annotations are drawn from the same distribution,
we can infer the statistics and therefore the consistency of the expert
annotations. If annotators choose not to annotate the instance, an
IoU of zero is assigned. Therefore, we expect outliers from the distribution
and will use the median IoU as a representative average metric. For
optimal annotations, the distribution will collapse to a Dirac distribution
with unit mean.

\subsection{Results}

\begin{figure}
\begin{centering}
\includegraphics[viewport=10bp 10bp 235bp 153bp,clip,width=1\columnwidth]{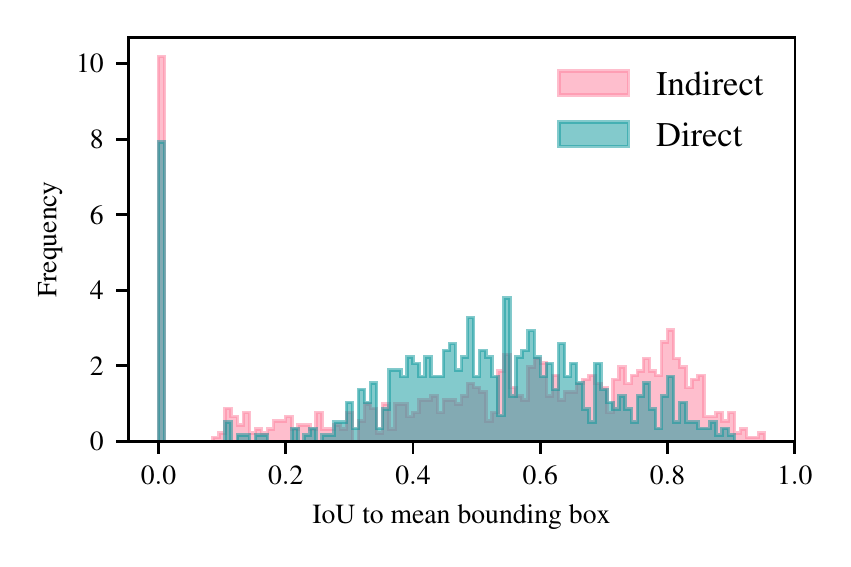}
\par\end{centering}
\caption{\label{fig:Comparison-of-Annotation}Histogram of IoU values cumulated
over all frames.}
\end{figure}
\begin{table}[b]
\caption{\label{tab:Classifier-architecture-Custom}Classifier architecture
of the custom model.}
\begin{tabular*}{1\columnwidth}{@{\extracolsep{\fill}}>{\centering}m{0.3\columnwidth}>{\centering}m{0.13\columnwidth}>{\centering}m{0.13\columnwidth}>{\centering}m{0.1\columnwidth}c}
\toprule 
Layer & Input channels & Output channels & Kernel size & Stride\tabularnewline
\midrule
Conv2D + ReLU & 1 & 32 & 5 & 1\tabularnewline
Conv2D + ReLU & 32 & 64 & 3 & 1\tabularnewline
MaxPool2D & 64 & 64 & 2 & 2\tabularnewline
BatchNorm2D & 64 & 64 & - & -\tabularnewline
Conv2D + ReLU & 64 & 64 & 5 & 1\tabularnewline
Conv2D + ReLU & 64 & 128 & 3 & 1\tabularnewline
MaxPool2D & 128 & 128 & 2 & 2\tabularnewline
BatchNorm2D & 128 & 128 & - & -\tabularnewline
Conv2D + ReLU & 128 & 128 & 5 & 1\tabularnewline
Conv2D + ReLU & 128 & 256 & 3 & 1\tabularnewline
AvgPool2D & 256 & 256 & 5 & 1\tabularnewline
BatchNorm2D & 256 & 256 & - & -\tabularnewline
Linear + ReLU + Dropout & 256 & 128 & - & -\tabularnewline
Linear + ReLU & 128 & 64 & - & -\tabularnewline
Linear + Sigmoid & 64 & 1 & - & -\tabularnewline
\bottomrule
\end{tabular*}
\end{table}
The results shown in \figref{Comparison-of-Annotation} support our
hypothesis, that a consistent, manual annotation with BBs is not possible.
Here, the abundance of missing annotations (mainly caused by different
starts and ending frames) shows, that even for a specified image region,
the experts could not agree on a consistent ground truth. Interestingly,
the direct annotations suffered from more inconsistency than the indirect
instance. We think, that this is mainly due to the big glare area
surrounding the instance, leading the experts to largely different
BB sizes. It is safe to say, that this inconsistency will cause problems
for systems learning from this ground truth data. Basically, each
new annotator introduces a domain shift into the annotations, which
makes it quite difficult to work with.

\end{document}